\documentclass{article}

\PassOptionsToPackage{numbers}{natbib}
\usepackage[preprint]{neurips_2025}

\usepackage[utf8]{inputenc} % allow utf-8 input
\usepackage[T1]{fontenc}    % use 8-bit T1 fonts
\usepackage[colorlinks=true, linkcolor=blue, citecolor=blue, urlcolor=blue]{hyperref}       % hyperlinks
\usepackage{url}            % simple URL typesetting
\usepackage{booktabs}       % professional-quality tables
\usepackage{amsfonts}       % blackboard math symbols
\usepackage{nicefrac}       % compact symbols for 1/2, etc.
\usepackage{microtype}      % microtypography
\usepackage{xcolor}         % colors
\usepackage{amsmath}
\usepackage{caption}
\usepackage{graphicx}
\usepackage{cleveref}

\title{Your Attention Matters: to Improve Model Robustness to Noise and Spurious
Correlations}

\author{
  Camilo Tamayo-Rousseau\thanks{Corresponding author. Email: camilo\_tamayo-rousseau@brown.edu} \\
  Brown University\\
  \And
  Yunjia Zhao \\
  Brown University \\
  \And
  Yiqun Zhang \\
  Brown University \\
  \And
  Randall Balestriero \\
  Brown University\\
}

\begin{document}

\maketitle

\begin{abstract}
  Self-attention mechanisms are foundational to Transformer architectures, supporting their impressive success in a wide range of tasks. While there are many self-attention variants, their robustness to noise and spurious correlations has not been well studied. This study evaluates Softmax, Sigmoid, Linear, Doubly Stochastic, and Cosine attention within Vision Transformers under different data corruption scenarios. Through testing across the CIFAR-10, CIFAR-100, and Imagenette datasets, we show that Doubly Stochastic attention is the most robust. It consistently outperformed the next best mechanism by $0.1\%-3.8\%$ in relative accuracy when training data, or both training and testing data, were corrupted. Our findings inform self-attention selection in contexts with imperfect data. The code used is available at \url{https://github.com/ctamayor/NeurIPS-Robustness-ViT}.
\end{abstract}
\begin{figure}[!h]
    \centering
    \includegraphics[height=8cm,width=\linewidth,keepaspectratio]{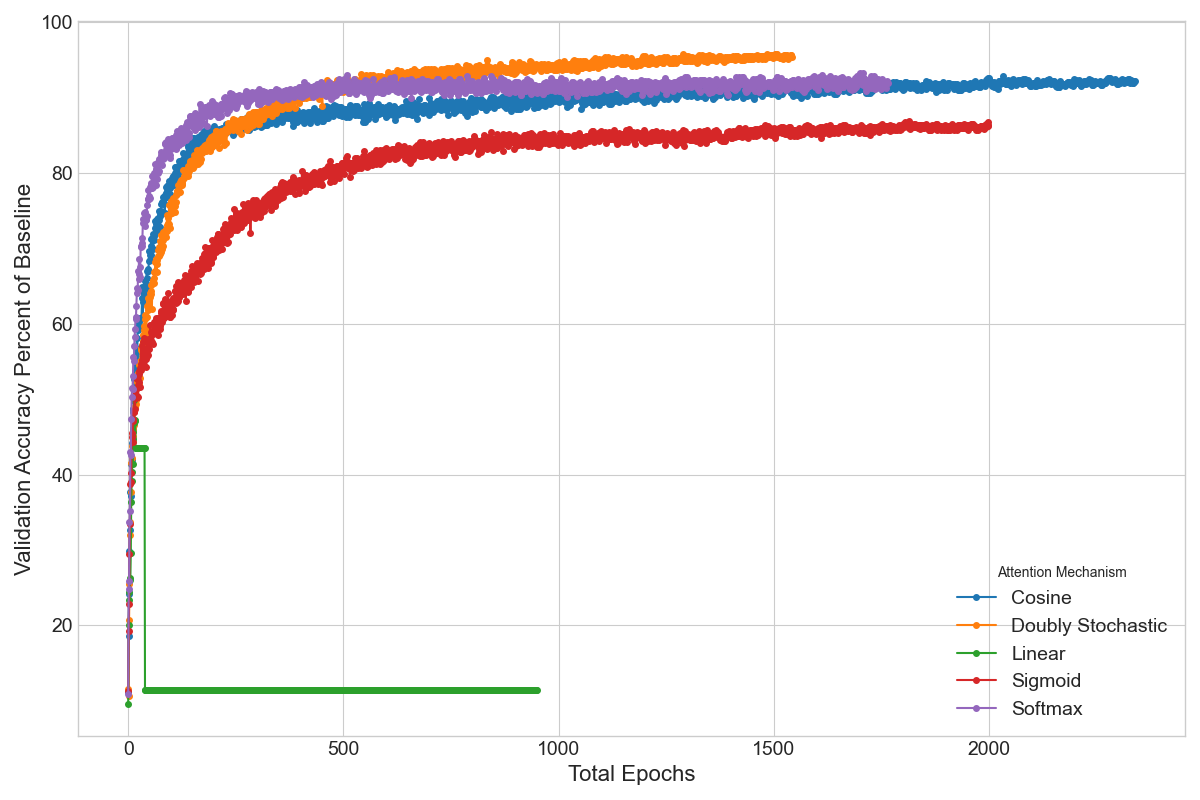}
    \caption{Relative validation accuracy (as a percentage of clean baseline) for Cosine (\textcolor{blue}{blue}), Doubly Stochastic (\textcolor{orange}{orange}), Linear (\textcolor{green}{green}), Sigmoid (\textcolor{red}{red}), and Softmax (\textcolor{violet}{purple}) attention on CIFAR-10 with corrupted training and testing data. We see clearly that Doubly Stochastic attention's performance degrades the least, as it consistently maintains the highest percentage of its original accuracy, confirming its superior resilience to corruption.}
    \label{fig:c10_absolute_training_curve}
\end{figure}

\section{Introduction}
Transformers \citep{vaswani2017attention} have become foundational in modern deep learning, achieving state-of-the-art results across natural language processing \citep{radford2018improving,devlin2019bert,yang2019xlnet}, computer vision \citep{dosovitskiy2020image,liu2021swin,carion2020end}, audio modeling \citep{gong2021ast,radford2023robust}, and multimodal tasks \citep{kim2021vilt,reed2022generalist,alayrac2022flamingo}. Their success is largely attributed to the self-attention mechanism, which models long-range dependencies by computing dynamic, content-based interactions between all input elements. This represents a shift from earlier architectures such as Convolutional Neural Networks (CNNs) \citep{lecun1998gradient, krizhevsky2012imagenet, he2016deep, simonyan2014very, szegedy2015going}, which rely on fixed, local receptive fields and parameter sharing to extract hierarchical features.

While CNNs have traditionally been the go-to models for vision tasks due to their inductive biases like locality and translation equivariance \citep{krizhevsky2012imagenet, lecun1998gradient}, recent work has shown that Vision Transformers (ViTs) \citep{dosovitskiy2020image} can match or even surpass CNNs when trained on large datasets. CNNs may exhibit some robustness to small perturbations due to their localized filters and smoothing behavior \citep{geirhos2018imagenet}, but they are still vulnerable to common corruptions such as noise, blur, and weather effects \citep{hendrycks2019benchmarking}. In contrast, ViTs use global self-attention mechanisms and lack built-in inductive biases, raising questions about their robustness under distribution shifts.

Robustness is increasingly vital in real-world applications, where inputs can be degraded due to sensor noise, occlusions, weather distortions, or even adversarial manipulations. While much of the robustness literature focuses on training-time interventions, architectural choices remain an underexplored but potentially complementary avenue. Existing approaches to improve robustness largely focus on modifying training procedures or data pipelines, such as data augmentation \citep{cubuk2019autoaugment,zhang2017mixup,devries2017improved}, adversarial training \citep{goodfellow2014explaining,madry2017towards}, or regularization techniques \citep{srivastava2014dropout, tarvainen2017mean,ioffe2015batch}. Although effective, these methods often introduce computational overhead or trade-offs between clean and corrupted data performance.

Notably, architectural innovations in self-attention mechanisms remain underexplored as pathways to robustness. While prior work has studied attention for improving efficiency \citep{wang2020linformer}\citep{mongaras2024cottention} or interpretability \citep{mrini2019rethinking}, its role in mitigating errors due to corruptions has not been systematically analyzed. Recent studies on Transformers for corrupted data \citep{bhojanapalli2021understanding}\citep{paul2022vision} primarily focus on comparing ViTs to CNNs rather than the architectural components. Some theoretical analysis has been done discussing kernel normalization strategies. When applied to the Gaussian kernel with heteroskedastic noise, doubly stochastic normalization was found to be more advantageous than row-stochastic and symmetric approaches \citep{landa2021doubly}. This leads one to wonder if this advantage holds in the context of Transformers and what the performance differences are between self-attentions.

To address this gap, we conduct a systematic study of how different attention mechanisms affect Transformer robustness to noisy and corrupted inputs. We compare five variants—Softmax attention, Sigmoid attention \citep{ramapuram2024theory}, Linear attention \citep{wang2020linformer, katharopoulos2020transformers, choromanski2020rethinking}, Doubly Stochastic attention \citep{sander2022sinkformers}, and Cosine attention \citep{mongaras2024cottention}—by integrating them into the Vision Transformer framework. Our key findings provide clear guidance on architectural selections for applications involving imperfect data:
\begin{itemize}
    \item Doubly Stochastic attention is the most robust choice. It consistently maintains the highest relative accuracy, particularly when both training and testing data are compromised. This makes it the ideal mechanism for ensuring stable performance in challenging, real-world conditions.
    \item Linear attention should be avoided in noisy environments as its performance can collapse entirely when trained on corrupted data. This instability makes it unsuitable for any application where data quality cannot be guaranteed.
    \item Robustness is best measured by relative performance, not just absolute accuracy. While standard Softmax attention often achieves high absolute accuracy, its performance degrades more significantly under corruption than Doubly Stochastic attention. The superior relative accuracy of Doubly Stochastic attention highlights its greater resilience and reliability when faced with imperfect data.
\end{itemize}

The code we used to train and evaluate our models is available at \url{https://github.com/ctamayor/NeurIPS-Robustness-ViT}.

\section{Attention Mechanism Choice is a Critical Factor for Model Robustness}
This section details the experimental framework used to evaluate the robustness of different attention mechanisms and presents the results of our analysis. The first part describes the setup, including the model architectures, datasets, and corruption methods. The second part discusses the performance outcomes, focusing on how each attention variant handles data corruption.

\subsection{Experimental Setup}
\label{sec:exp_setup}
We integrated the five attention mechanisms into a Vision Transformer (ViT) backbone. Vision Transformers adapt the Transformer architecture to process visual information. ViTs treat an input image as a sequence of non-overlapping patches which are flattened and then projected onto a higher dimensional space using a learnable embedding matrix. This is combined with positional embeddings and fed into a sequence of encoder layers. The layers apply self-attention mechanisms and feed-forward neural networks to arrive at the desired output.

Central to the ViT is the self-attention mechanism, which computes pairwise interactions between all tokens in a sequence. For queries $Q\in \mathbb{R}^{n \times d}$, keys $K\in\mathbb{R}^{n\times d}$, and values $V\in\mathbb{R}^{n\times d}$, the attention output is computed as:
\begin{equation}
    \text{Attention}(Q,K,V;\phi)=\phi\left(QK^T\right)V
    \label{eq:attention}
\end{equation}
where $\phi$ is the activation function. The term $QK^T\in R^{n\times n}$ measures the dot-product similarity between each query-key pair, and the output $\phi\left(QK^T\right)V$ is a weighted sum of $V$ where a value has more weight if it's key has a larger similarity with the query. To improve representational power, ViTs use an extension of this attention design called Multi-Head attention. Instead of having one all-encompassing map, the queries, keys, and values are split into $h$ parallel heads. Each head independently computes attention, and the outputs are combined to form the final result, enabling the model to attend to diverse patterns across different subspaces. 

To ensure fair comparisons, we independently tuned hyperparameters for each mechanism on clean data including architectural (e.g. number of heads and hidden dimension), optimization (e.g. batch size and learning rate), and attention specific settings. Attention mechanisms are central to our study so we will now clarify the specifics of each attention mechanism within the Transformer framework.

\textbf{Softmax attention} is when the activation function $\phi$ is the matrix softmax map that applies softmax row-wise:
\begin{equation}
    \small
    \text{softmax}\left(\left[
    \begin{array}{c@{\hskip 2pt}c@{\hskip 2pt}c}
        x_{11} & \cdots & x_{1n} \\
        \vdots & \vdots & \vdots \\
        x_{n1} & \cdots & x_{nn}
    \end{array}\right]
    \right) =\left[
    \small
    \begin{array}{c@{\hskip 2pt}c@{\hskip 2pt}c}
        \frac{e^{x_{11}}}{\sum_{j=1}^n e^{x_{1j}}} & \cdots & \frac{e^{x_{1n}}}{\sum_{j=1}^n e^{x_{1j}}} \\
        \vdots & \vdots & \vdots \\
        \frac{e^{x_{n1}}}{\sum_{j=1}^n e^{x_{nj}}} & \cdots & \frac{e^{x_{nn}}}{\sum_{j=1}^n e^{x_{nj}}}
    \end{array}\right]
    \label{eq:softmax_expansion}
\end{equation}

To prevent gradients from being too small, the factor $\frac{1}{\sqrt{d}}$ is applied to the matrix prior to the softmax computation. This gives us:
\begin{equation}
    \text{SoftmaxAttn}(Q,K,V)=\text{softmax}\left(\frac{QK^T}{\sqrt{d}}\right)V
    \label{eq:softmax}
\end{equation}

\textbf{Sigmoid attention}
is when the activation function becomes the sigmoid function
\begin{equation}
    \sigma(u+b) = (1+e^{-(u+b)})^{-1}
    \label{eq:sigmoid function}
\end{equation}
Here, a bias term $b\in \mathbb{R}$ is added as a hyperparameter. In the full Sigmoid attention definition, the scaling term is applied as in Softmax attention:
\begin{equation}
    \text{SigmoidAttn}(Q,K,V)=\sigma\left(\frac{QK^T}{\sqrt{d}}\right)V
    \label{eq:sigmoid attention}
\end{equation}
$\sigma$ is applied to the input matrix element-wise. We also added a LayerScale layer \citep{touvron2021going} as was done in the original Sigmoid self-attention paper \citep{ramapuram2024theory}.

\textbf{Linear attention} reduces computation time by applying an activation function that has a linear time complexity. We apply the linear feature map $\psi(X) = \text{ELU}(X)+1$ which leads us to the definition:
\begin{equation} \text{LinearAttn}(Q, K, V) = \left( \frac{\psi(Q) (K V)}{\psi(Q) K^T + \epsilon} \right) \label{eq:linear attention} \end{equation}
The $\epsilon$ term is a small constant for numerical stability.

\textbf{Doubly Stochastic attention} imposes the constraint that the rows and columns of the attention matrix must each sum to one. This is accomplished using Sinkhorn's algorithm which iterates over a matrix $C$, starting from $K^0=\text{exp}(C)$:\\
\begin{equation} 
    K^{l+1}=\left\{ \begin{array}{cc}
        N_R(K^l) & \text{if $l$ is even} \\
        N_C(K^l) & \text{if $l$ is odd}
    \end{array}\right.
    \label{eq:sinkorns_algorithm}
\end{equation}
where $N_R$ and $N_C$ are the row-wise and column-wise normalizations: $(N_R(K))_{i,j}=\frac{K_{i,j}}{\sum_{l=1}^nK_{i,l}}$ and $(N_C(K))_{i,j}=\frac{K_{i,j}}{\sum_{l=1}^nK_{i,j}}$.

For better numerical stability, the algorithm transitions into the log domain where $u$ and $v$ are introduced to enforce the row and column stochasticity. To perform iterative updates, a cost function is used:
\begin{equation}
    M_{ij} = \frac{-c_{ij}+u_i+v_j}{\epsilon}
    \label{eq:modified_cost}
\end{equation}
where the additional hyperparameter $\epsilon$ acts as a step-size. The updates for $u$ and $v$ become:
\begin{equation}
    \begin{aligned}
        u_i^{l+1}=u_i^l + \epsilon(\log \mu_i - \log\sum_j\exp(M_{ij}))\\
    v_j^{l+1}=v_j^l + \epsilon(\log \nu_j - \log\sum_i\exp(M_{ij}))
    \end{aligned}
    \label{eq:uv_updates}
\end{equation}
with $\mu$ and $\nu$ being the desired row and column sums respectively. The number of iterations (another hyperparameter) is capped to balance convergence quality with computational cost.

So, Doubly Stochastic attention is computed via the following equation:
\begin{equation} \text{DoublyStochasticAttn}(Q, K, V) = \text{Sinkhorn} \left( \frac{QK^T}{\sqrt{d}} \right) V
\label{eq:doubly stochastic attention}
\end{equation}
The scaling factor $\sqrt{d}$ where $d=\frac{\text{features}}{\text{heads}}$ improves computation stability.

\textbf{Cosine attention} can be written as:
\begin{equation} \text{CosineAttn}(Q, K, V) = \frac{\text{cos}(Q, K)}{n^{\sigma(m)}} V \label{eq:cosine_attention} \end{equation}
where $\cos(Q,K)=\frac{QK^T}{\|Q\|\|K\|}$ computes cosine similarity between queries and keys, and $\sigma(m)$ is a learned stabilization parameter applied via the sigmoid function.

Experiments spanned the CIFAR-10 \citep{krizhevsky2009learning}, CIFAR-100 \citep{krizhevsky2009learning}, and Imagenette \citep{Howard_Imagenette_2019} datasets to evaluate robustness across resolutions (32 x 32 to 224 x 224) and class counts (10 to 100) as shown below.
\begin{itemize}
    \setlength\itemindent{2em}
    \item CIFAR-10 (C10): 32 x 32 resolution and 10 classes
    \item CIFAR-100 (C100): 32 x 32 resolution and 100 classes
    \item Imagenette: rescaled to 224 x 224 resolution and 10 classes
\end{itemize}
All images were normalized using mean and standard deviation values specific to their dataset. 

During training, we applied AutoAugment policies \citep{cubuk2019autoaugment}.
Initial experiments with gaussian noise lacked significant performance variation across attention mechanisms, leading to a shift to fog corruption which better reflects real-world degradation. Fog severity was tuned to levels causing noticeable validation accuracy drops compared to clean data. Corruption was applied in three settings:
\begin{itemize}
\item Clean Training + Corrupted Testing: Models trained on clean data and tested on fog.
\item Corrupted Training + Corrupted Testing: Models trained on fog and tested on fog.
\item Corrupted Training + Clean Testing: Models trained on fog and tested on clean images.
\end{itemize}

\subsection{Doubly Stochastic is the Most Robust Attention Mechanism}
We evaluate the attention performance under the four data corruption scenarios: no corruption, corruption applied only during training, only during testing, and during both training and testing. To offer a comprehensive robustness assessment, we analyze both absolute accuracies—reflecting raw performance—and relative accuracies, which measure performance against a clean baseline. This allows us to distinguish between overall predictive strength and resilience to corruption.

In evaluating model robustness to data corruption, both absolute and relative accuracies offer valuable but distinct insights. Absolute accuracy refers to a model’s raw performance under each corruption scenario—whether on clean data, corrupted training data, corrupted test data, or both. These scores are critical for understanding a model’s overall predictive capability in each condition. For instance, a model that achieves high accuracy even when exposed to corrupted inputs may be both robust and inherently performant. Absolute accuracies also help assess whether a model is viable for real-world deployment, where minimum performance thresholds may be required regardless of conditions.

However, absolute accuracy alone does not tell the full story, especially when comparing models with different baselines. This is where relative accuracy becomes important. Relative accuracy quantifies performance degradation by comparing corrupted-case performance to the clean (no-corruption) baseline, which we express as a percentage of the baseline accuracy. This normalization allows for a fair comparison of robustness across models with differing clean accuracies by focusing on how much performance is lost, not just the final score. So, we analyze both the absolute and relative accuracies from our experiments. 

\begin{figure}[!h]
    \centering
    \begin{minipage}[t]{0.48\textwidth} 
        \centering
        \includegraphics[height=7cm,width=\linewidth,keepaspectratio]{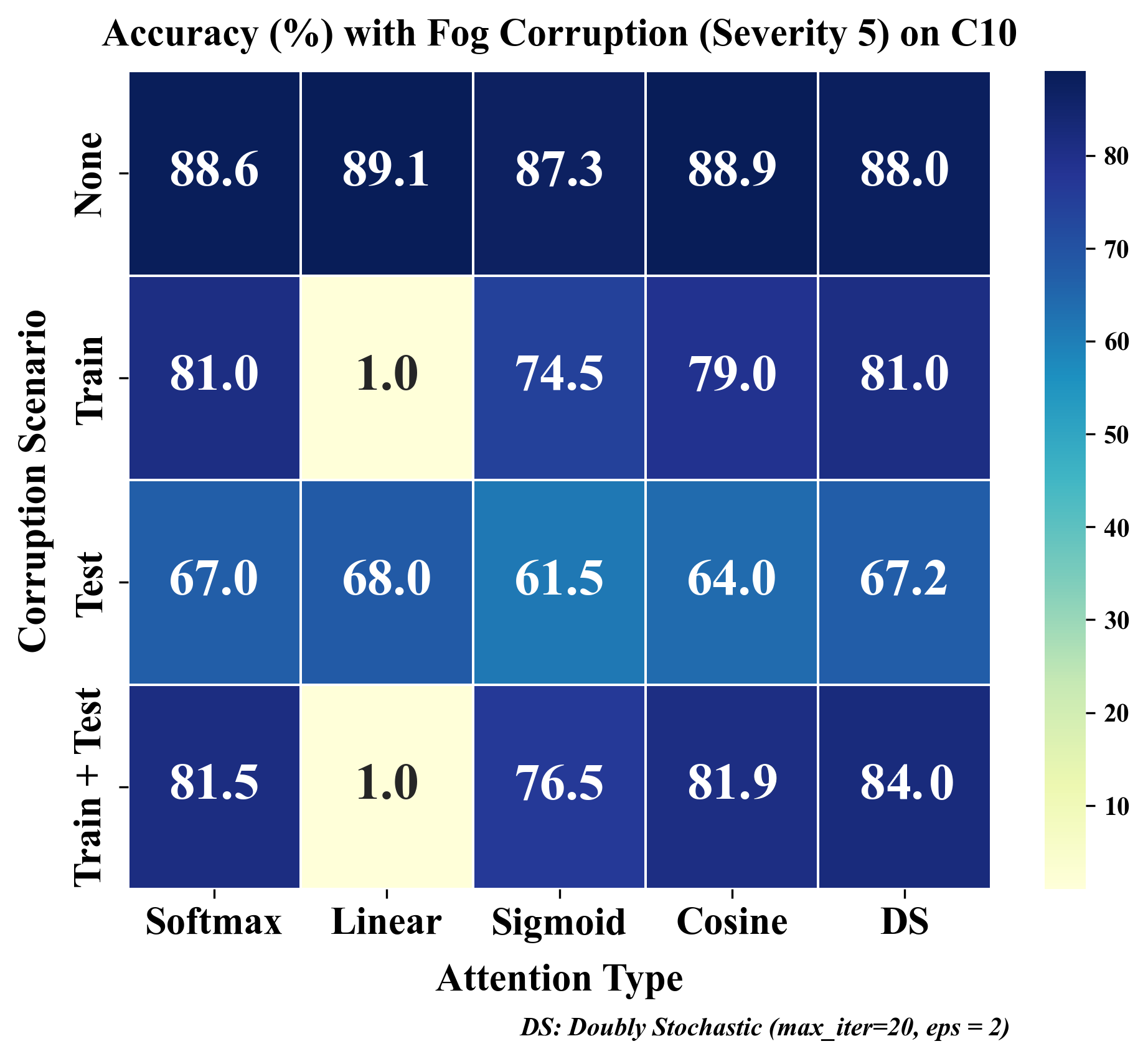}
        \caption{Comparison of absolute validation accuracy (\%) of five attention mechanisms—Softmax, Linear, Sigmoid, Cosine, and Doubly Stochastic—on the CIFAR-10 dataset under four corruption settings. Doubly Stochastic attention achieves the highest absolute accuracy on CIFAR-10 when both training and test data are corrupted, while Softmax performs best on clean data and Linear attention collapses in the settings involving corrupted training data due to gradient instability. See \cref{fig:fog_radar_c10} for a depiction of relative accuracy on CIFAR-10.}
        \label{fig:fog_heatmap_c10}
    \end{minipage}
    \hfill
    \begin{minipage}[t]{0.48\textwidth} 
    \centering
    \includegraphics[width=1\textwidth]{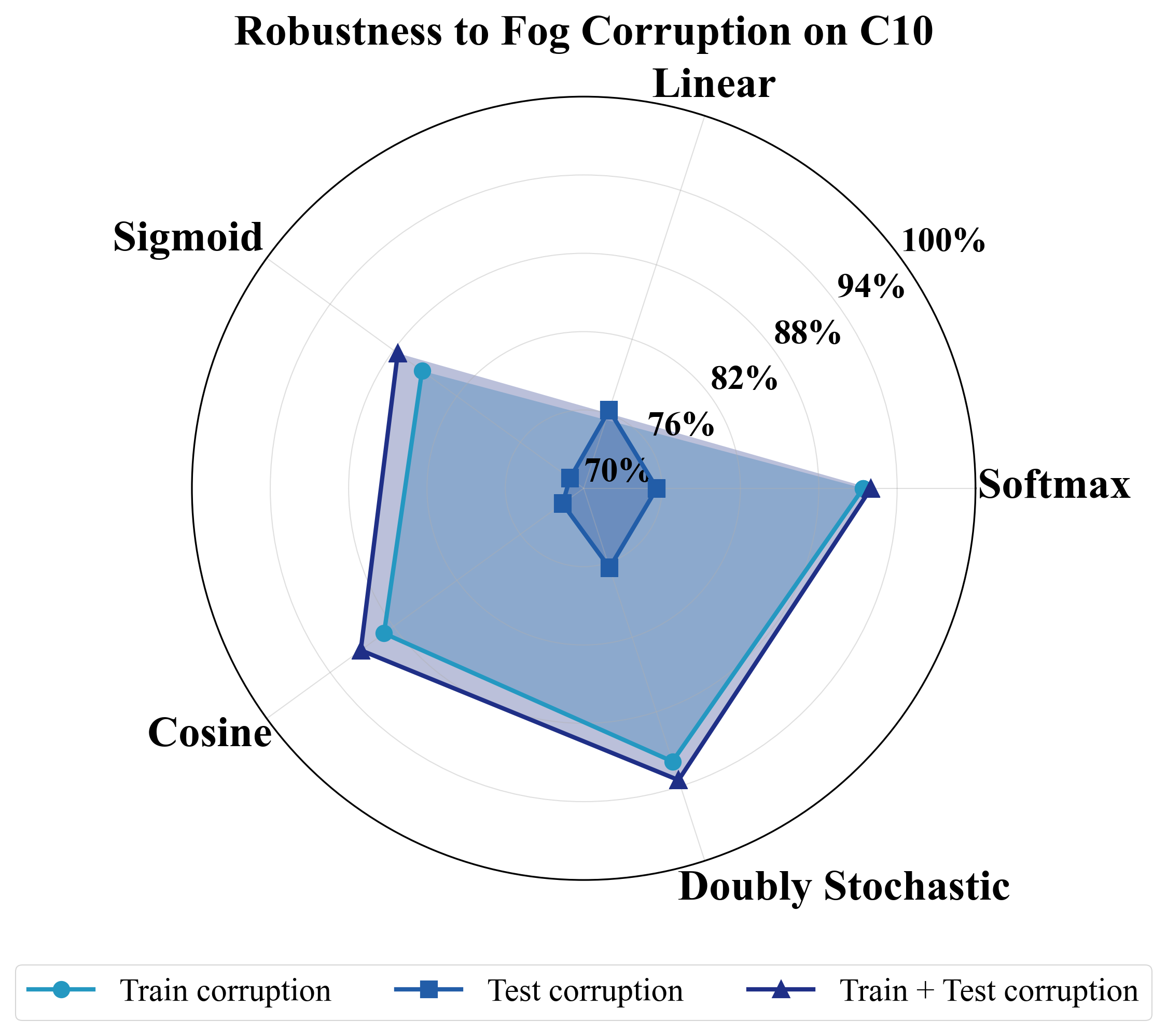}
    \caption{Depiction of relative validation accuracy (\%) of five attention mechanisms—Softmax, Linear, Sigmoid, Cosine, and Doubly Stochastic—on the CIFAR-10 dataset under four corruption settings. Doubly Stochastic attention demonstrates superior robustness to fog corruption on CIFAR-10, with Linear attention showing notable degradation when training data is corrupted (due to gradient instability). See \cref{fig:fog_heatmap_c10} for a visualization of absolute accuracy on CIFAR-10.}
    \label{fig:fog_radar_c10}
\end{minipage}
\end{figure}

\begin{table*}[!h]
\centering
\caption{Performance of five attention mechanisms (\textbf{columns}) on four distinct CIFAR-10 corruption settings (\textbf{rows}): no corruptions, train corruption only, test corruption only, and both train and test corruption. We see that for absolute accuracy, no attention is superior while for accuracy relative to the clean baseline (in parenthesis), Doubly Stochastic leads in every corruption setting.}
\label{tab:cifar10_twocol}
\resizebox{\textwidth}{!}{%
\begin{tabular}{lccccc}
\toprule
\textbf{Condition} & \textbf{Softmax} & \textbf{Linear} & \textbf{Sigmoid} & \textbf{Cosine} & \textbf{Doubly Stochastic} \\
& & & & & \textit{(max\_iter=20, eps=2)} \\
\midrule
\textbf{No corruption}
& 88.6\% & 89.1\% & 87.3\% & 88.9\% & 88.0\% \\

\textbf{Train corruption}
& 81.0\% (91.4\%)
& 1.0\% (1.1\%)
& 74.5\% (85.3\%)
& 79.0\% (88.9\%)
& 81.0\% (\textbf{92.0\%}) \\

\textbf{Test corruption}
& 67.0\% (75.6\%)
& 68.0\% (76.3\%)
& 61.5\% (71.3\%)
& 64.0\% (72.0\%)
& 67.2\% (\textbf{76.4\%}) \\

\textbf{Train + Test corruption}
& 81.5\% (92.0\%)
& 1.0\% (1.1\%)
& 76.5\% (87.6\%)
& 81.9\% (92.5\%)
& 84.0\% (\textbf{95.5\%}) \\
\bottomrule
\end{tabular}
}
\end{table*}
Experimental results for the CIFAR-10 dataset are presented in \cref{tab:cifar10_twocol} and visualized in \cref{fig:fog_heatmap_c10} and \cref{fig:fog_radar_c10}. In terms of absolute accuracy, Softmax and Doubly Stochastic attention yielded nearly identical performance under both training-only and testing-only corruption scenarios. However, Doubly Stochastic attention achieved a higher absolute accuracy ($2.1$\% improvement over the next best attention) when corruption was present during both training and testing.

Examining relative accuracy reveals a clearer distinction in robustness. Across all three corruption settings, Doubly Stochastic attention consistently maintained a higher percentage of its clean data accuracy compared to the other mechanisms. With corruption only applied to training images, Doubly Stochastic had a relative accuracy of $92.0\%$ while the next highest was Softmax with a relative accuracy of $91.4\%$. With test corruption, Doubly Stochastic's relative accuracy was $76.4\%$ with Linear attention following with $76.3\%$. When both train and test corruption were applied, Doubly Stochastic had an accuracy of $95.5\%$ followed by Cosine with $92.5\%$.

These results suggest that Doubly Stochastic attention consistently demonstrates the most resilience to corruption, even when its absolute performance is not the highest. Although Softmax attention performs comparably in some cases, especially under single-source corruption, Doubly Stochastic maintains the highest relative accuracies across all three corrupted settings, suggesting it is less sensitive to the presence of noise in either training or testing data.
Linear attention, on the other hand, performs adequately under test-time corruption but completely collapses when training data is corrupted—highlighting a severe brittleness during learning.
\begin{table*}[!h]
\centering
\caption{Performance of five attention mechanisms (\textbf{columns}) on four distinct CIFAR-100 corruption settings (\textbf{rows}): no corruptions, train corruption only, test corruption only, and both train and test corruption. We see that for absolute accuracy, Softmax outperforms the rest while for accuracy relative to the clean baseline (in parenthesis), Doubly Stochastic leads in train-only corruption as well as in train and test corruption.}
\label{tab:cifar100_twocol}
\resizebox{\textwidth}{!}{%
\begin{tabular}{lccccc}
\toprule
\textbf{Condition} & \textbf{Softmax} & \textbf{Linear} & \textbf{Sigmoid} & \textbf{Cosine} & \textbf{Doubly Stochastic} \\
& & & & & \textit{(max\_iter=6, eps=1.0)} \\
\midrule
\textbf{No corruption}
& 65\% & 55\% & 57.5\% & 62\% & 59.2\% \\

\textbf{Train corruption}
& 49\% (75.4\%)
& 1\% (1.8\%)
& 38.5\% (67.0\%)
& 48\% (77.4\%)
& 46\% (\textbf{77.7\%}) \\

\textbf{Test corruption}
& 33.5\% (51.5\%)
& 21.5\% (39.1\%)
& 24\% (41.7\%)
& 32\% (\textbf{51.6\%})
& 29\% (49.0\%) \\

\textbf{Train + Test corruption}
& 52\% (80.0\%)
& 1\% (1.8\%)
& 40\% (70.0\%)
& 48.4\% (78.1\%)
& 48.5\% (\textbf{81.9\%}) \\
\bottomrule
\end{tabular}
}
\end{table*}

\begin{figure}[!h]
    \centering
    \begin{minipage}[t]{0.48\textwidth} 
    \centering
    \includegraphics[height=7cm,width=\linewidth,keepaspectratio]{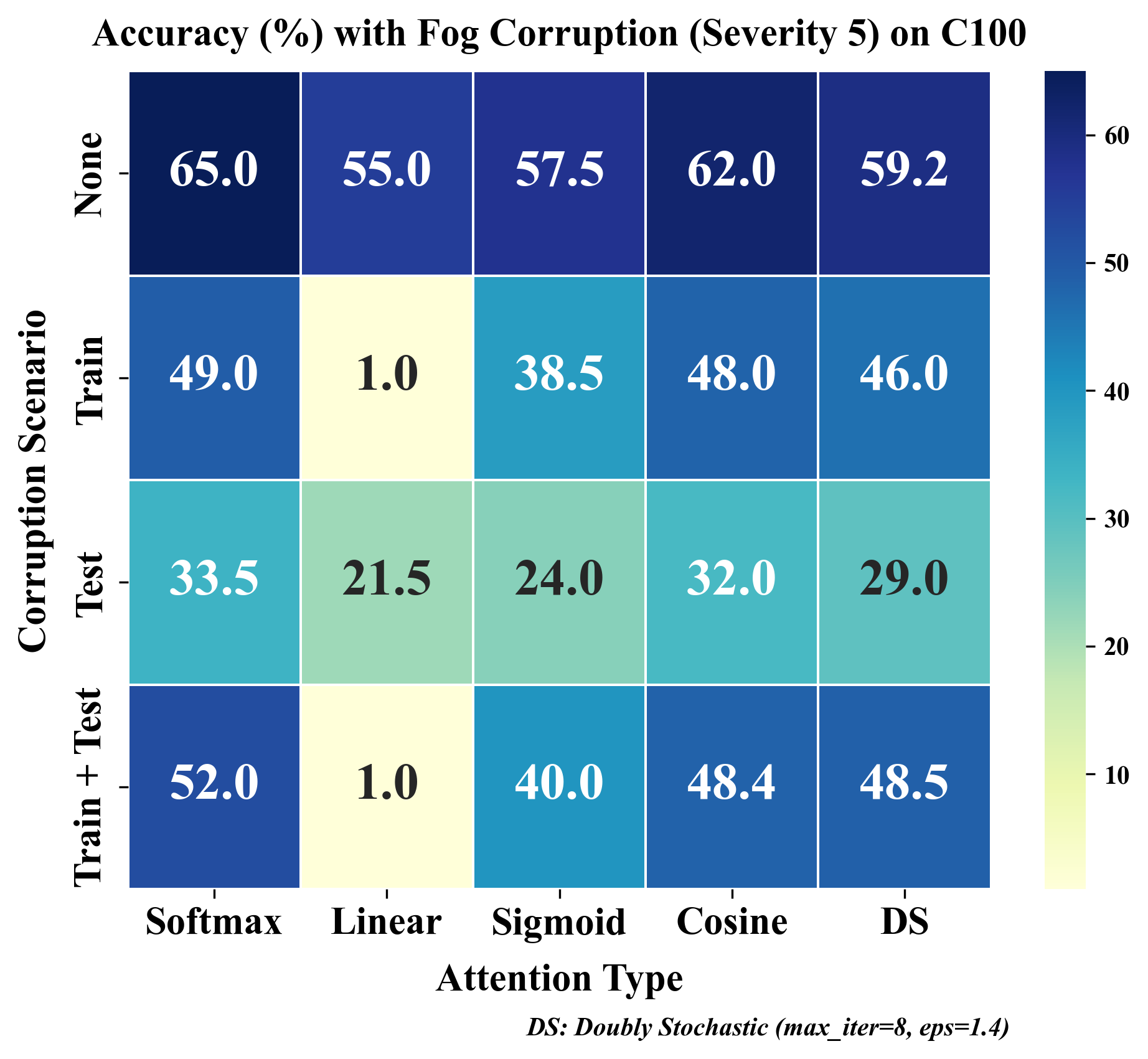}
        \caption{Analysis of the absolute validation accuracy (\%) of five attention mechanisms (\textbf{x-axis}) on the CIFAR-100 dataset under four distinct corruption scenarios (\textbf{y-axis}). Softmax attention achieves the highest absolute accuracy on CIFAR-100 in all corruption settings. Linear attention collapses in the settings involving corrupted training data due to gradient instability. See \cref{fig:fog_radar_c100} for a visualization of relative accuracy on CIFAR-100.}
        \label{fig:fog_heatmap_c100}
    \end{minipage}
    \hfill
    \begin{minipage}[t]{0.48\textwidth}
    \centering \includegraphics[height=7cm,width=\linewidth,keepaspectratio]{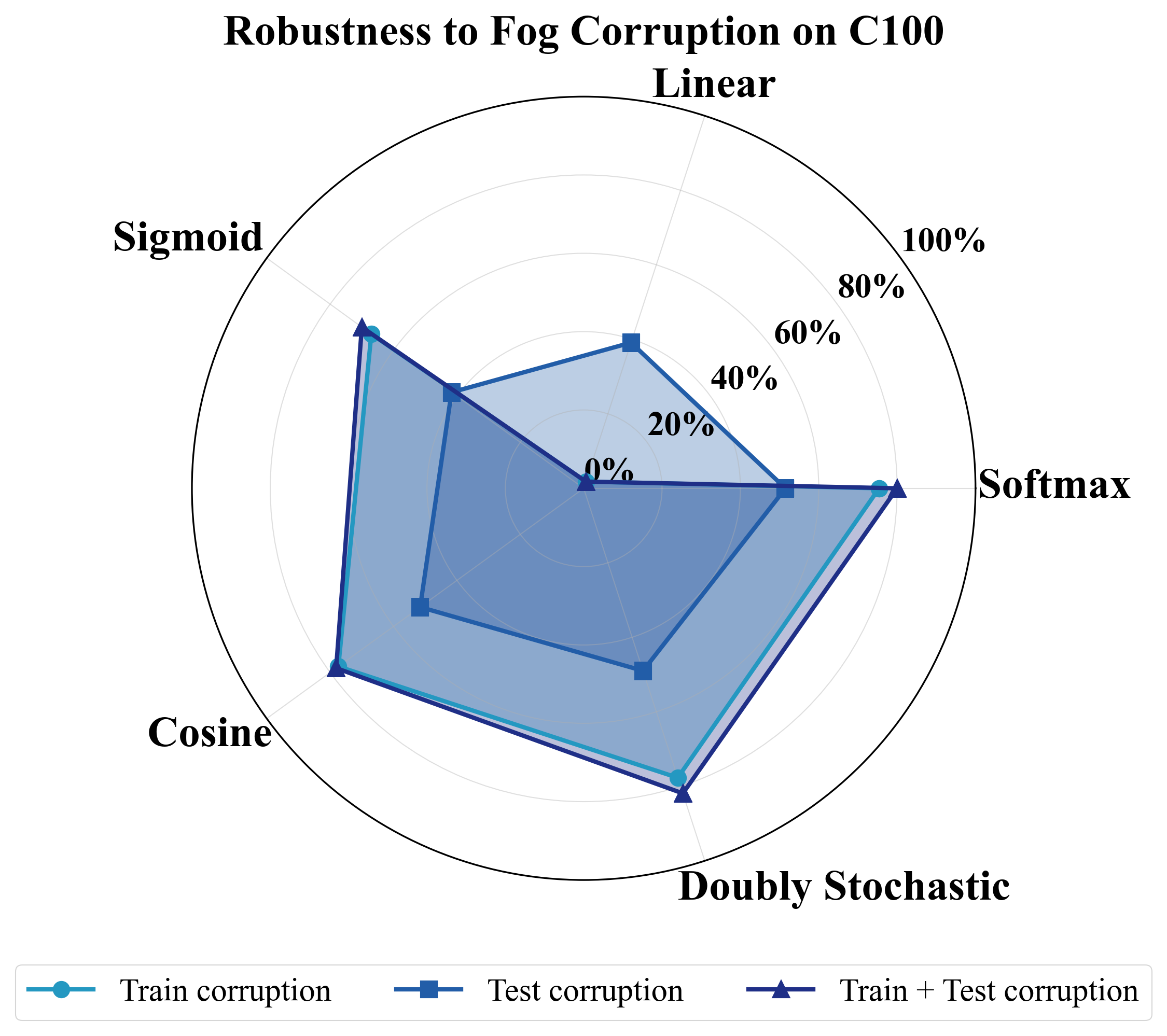}
    \caption{Visualization of relative validation accuracy (\%) of five attention mechanisms—Softmax, Linear, Sigmoid, Cosine, and Doubly Stochastic—on the CIFAR-100 dataset under four corruption settings. Doubly Stochastic attention demonstrates superior robustness to fog corruption on CIFAR-100 in settings involving corrupted training data, while Softmax is best on test-only corruption. See \cref{fig:fog_heatmap_c100} for an analysis of absolute validation accuracy on CIFAR-100.}
    \label{fig:fog_radar_c100}
\end{minipage}
\end{figure}

\Cref{tab:cifar100_twocol}, along with \Cref{fig:fog_heatmap_c100} and \Cref{fig:fog_radar_c100} present the findings for the CIFAR-100 dataset. Softmax attention achieved the highest absolute accuracy across all corruption scenarios. However, in terms of relative accuracy, Doubly Stochastic attention (77.7\%) had the highest accuracy on train corruption, narrowly surpassing cosine attention (77.4\%). When corruption was applied only to test images, Cosine demonstrated the highest relative accuracy (51.6\%) followed by Softmax (51.5\%) and then Doubly Stochastic (49.0\%). When both training and test sets were corrupted, Doubly Stochastic showed a more pronounced advantage over the others, achieving an 81.9\% relative accuracy followed by Softmax at 80.0\% relative accuracy.

These results largely align with the C10 findings. They suggest that Doubly Stochastic has a significant robustness advantage under combined training and testing corruptions, and a marginal benefit when only train data is corrupted. Unlike with C10, Doubly Stochastic does not outperform in the test corruption setting on C100.
\begin{table*}[!h]
\centering
\caption{Comparison of five attention mechanisms (\textbf{columns}) on the Imagenette dataset under four distinct corruption settings (\textbf{rows}): no corruptions, train corruption only, test corruption only, and both train and test corruption. With every corruption scenario, Doubly Stochastic attention retains the highest percentage of its clean accuracy (relative accuracy shown in parenthesis), indicating more robustness to corruption.}
\label{tab:imagenette_twocol}
\resizebox{\textwidth}{!}{%
\begin{tabular}{lccccc}
\toprule
\textbf{Condition} & \textbf{Softmax} & \textbf{Linear} & \textbf{Sigmoid} & \textbf{Cosine} & \textbf{Doubly Stochastic} \\
& & & & & \textit{(max\_iter=6, eps=.2)} \\
\midrule
\textbf{No corruption}
& 77\% & 76\% & 68\% & 75.2\% & 75.3\% \\

\textbf{Train corruption}
& 70\% (90.9\%)
& 67.5\% (88.8\%)
& 60.5\% (89.0\%)
& 66\% (87.7\%)
& 68.5\% (\textbf{91.0\%}) \\

\textbf{Test corruption}
& 63\% (81.8\%)
& 58\% (76.3\%)
& 49\% (72.1\%)
& 56\% (74.5\%)
& 63.5\% (\textbf{84.3\%}) \\

\textbf{Train + Test corruption}
& 73\% (94.8\%)
& 73\% (96.1\%)
& 63\% (92.6\%)
& 69\% (91.8\%)
& 75.2\% (\textbf{99.9\%}) \\
\bottomrule
\end{tabular}
}
\end{table*}

\begin{figure}[!h]
    \centering
    \begin{minipage}[t]{0.48\textwidth} 
        \centering
        \includegraphics[height=7cm,width=\linewidth,keepaspectratio]{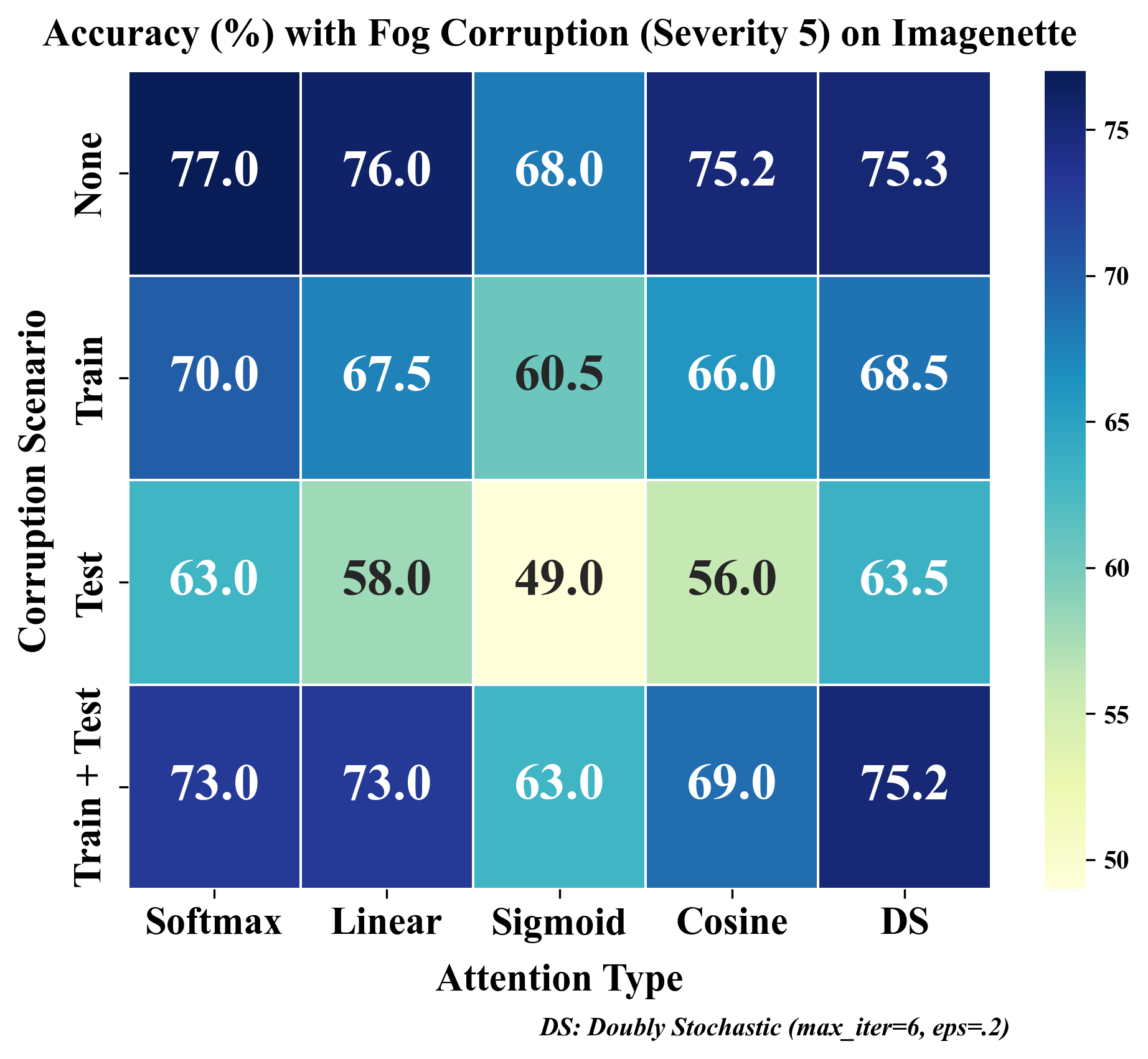}
        \caption{Analysis of the absolute validation accuracy (\%) of five attention mechanisms (\textbf{x-axis}) on the Imagenette dataset under four distinct corruption scenarios (\textbf{y-axis}). Softmax attention achieves the highest absolute accuracy on clean data and train-only corruption while Doubly Stochastic attention performed best on test-only corruption and when train and test data were corrupted. Linear attention collapses in the settings involving corrupted training data due to gradient instability. See \cref{fig:fog_radar_imagenette} for a visualization of relative accuracy on Imagenette.}
        \label{fig:fog_heatmap_imagenette}
    \end{minipage}
    \hfill
    \begin{minipage}[t]{0.48\textwidth} 
    \centering
    \includegraphics[height=7cm,width=\linewidth,keepaspectratio]{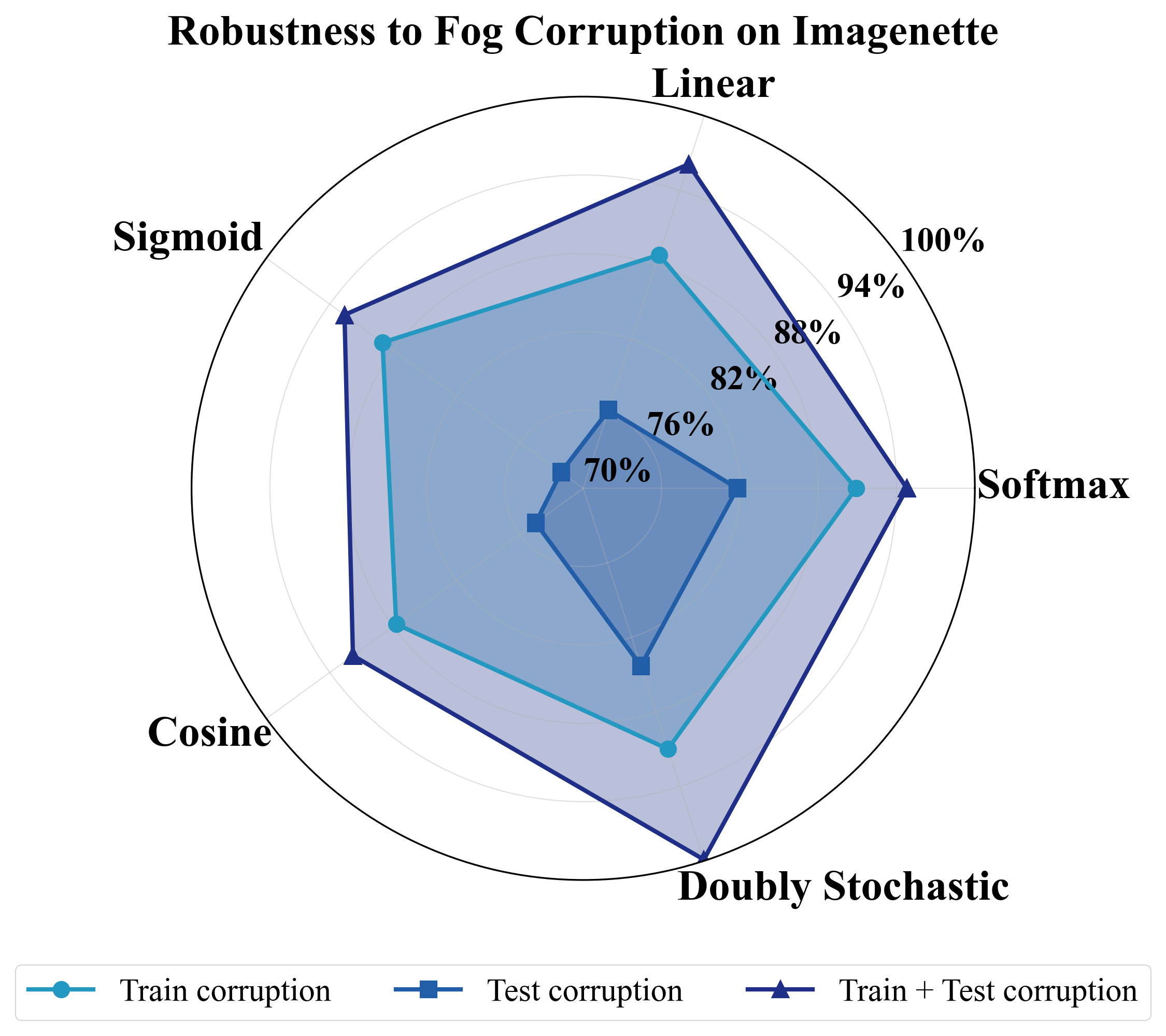}
    \caption{Visualization of relative validation accuracy (\%) of five attention mechanisms—Softmax, Linear, Sigmoid, Cosine, and Doubly Stochastic—on the Imagenette dataset under four corruption settings. Doubly Stochastic attention demonstrates superior robustness to fog corruption in all scenarios. See \cref{fig:fog_heatmap_imagenette} for a comparison of absolute accuracy on Imagenette.}
    \label{fig:fog_radar_imagenette}
\end{minipage}
\end{figure}

For the Imagenette dataset, the experiment results are detailed \Cref{tab:imagenette_twocol} and illustrated in \Cref{fig:fog_heatmap_imagenette} and \Cref{fig:fog_radar_imagenette}. Softmax attention achieved the highest absolute accuracy on clean data and under training-only corruption. Conversely, Doubly Stochastic attention led in absolute accuracy for test-only corruption and when corruption affected both training and testing data. Regarding relative accuracy, Doubly Stochastic attention again showed a marginal lead in the training corruption scenario. Doubly Stochastic demonstrated significantly greater robustness in the test-only setting and, notably, in the combined training train and test scenario where it remained almost entirely unaffected by perturbations.

These Imagenette results further underscore the superior robustness of Doubly Stochastic attention, particularly when facing combined training and testing data corruption. Furthermore, the trend of slightly improved relative accuracy with corrupted training images for Doubly Stochastic attention continues to be evident.
\section{Conclusion}
\label{sec: conclusion}
In this work, we demonstrated that the choice of attention critically influences model resilience. Our experiments revealed the superior relative robustness of Doubly Stochastic attention, particularly when training data was corrupted, and most notably when both training and testing data were compromised. This resilience highlights the potential benefits of its inherent row and column normalization constraints in stabilizing performance under challenging conditions.

These findings have significant implications for real-world applications where data corruption is inevitable. Our results emphasize that attention mechanisms should be selected not only for their performance on clean-data, but also for their ability to withstand noise and maintain reliability when faced with spurious correlations.

While our study provides valuable insights, we acknowledge certain limitations. The investigation primarily centered on fog corruption, chosen for its efficacy in differentiating model performance. The generalization of the observed robustness hierarchies to a broader spectrum of data degradations, such as various types of image noise, blur, or other common corruptions, warrants further dedicated study. Similarly, our experiments were conducted using a specific set of vision datasets (CIFAR-10, CIFAR-100, and Imagenette) and a standardized Vision Transformer backbone. Consequently, the direct applicability of our findings to vastly different data domains, tasks beyond image classification, or significantly altered transformer architectures remains an open question.

In addition, due to time and resource constraints, our experiments were conducted using a single random seed per configuration. As a result, we do not report error bars or statistical significance measures. We acknowledge this as a limitation, as performance variability across runs could affect the robustness conclusions. However, we note that the trends were consistent across all datasets and corruption settings, which we believe supports the reliability of the observed patterns. Future work will incorporate multiple runs to provide a more rigorous statistical analysis.

Future research should aim to address these aspects. Extending this comparative analysis to a wider array of corruption types would provide a more comprehensive understanding. Evaluating these attention mechanisms on a more diverse range of datasets, encompassing different modalities and tasks, as well as exploring their behavior within different transformer architectures and scales, will be crucial for assessing the broader generalizability of these findings. A thorough investigation into the trade-offs between robustness gains and the computational costs associated with each attention mechanism would also offer critical practical guidance for model selection.

Ultimately, this study underscores that the choice of attention mechanism is not a trivial detail but a significant factor in a model's ability to adapt to and perform reliably with the imperfect data characteristic of real-world applications. For robustness, your attention, indeed, matters.

{
    \small
    \bibliographystyle{plainnat}
    \bibliography{main}
}

\appendix
\section{Compute Resources}
All experiments were conducted on a high-performance computing (HPC) cluster. Each job was run on a compute node with access to 48 CPU cores, 384 GB of RAM, and up to 8 standard NVIDIA GPUs (Quadro RTX or equivalent). Individual training runs for each attention mechanism typically took approximately 10 hours. Hyperparameter tuning was conducted separately for each attention variant on each dataset, which required significantly more runs and compute time. We estimate that the full experimental study consumed roughly 80 GPU-days in total. The increased cost was primarily due to tuning architectural parameters to ensure fair comparison across attention types.
\section{Training Details}
Training was performed from scratch on modified versions of the ViT-Base architecture. All models were trained using the Adam optimizer with a weight decay of $5e$-$5$. The learning rate was warmed up for $5$ epochs and then decayed using a cosine scheduler to a minimum learning rate of $1e$-$3$. The number of layers was also held constant at $7$ for all experiments.

Certain hyperparameters were modified based on variations in the attention mechanism and dataset. The hyperparameters used in the experiments are shown below.

\begin{table}[h!]
\label{tab:hyperparam_variations}
\centering
\begin{tabular}{c|cccccc}
    \toprule
    \textbf{Attention} & Softmax & Softmax & Softmax & 
    Sigmoid & Sigmoid & Sigmoid\\
    \textbf{Dataset} & C10 & C100 & Imagenette & 
    C10 & C100 & Imagenette \\
    \textbf{Init LR} & 1e-3 & 1e-3 & 1e-3 & 
    3e-4 & 3e-4 & 5e-5
    \\
    \textbf{Batch Size} & 128 & 128 & 128 &
    64 & 128 & 128 
    \\
    \textbf{Dropout} & 0.0 & 0.0 & 0.0 &
    0.1 & 0.0 & 0.0 
    \\
    \textbf{Heads} & 12 & 12 & 12 &
    12 & 12 & 12 
    \\
    \textbf{Hidden} & 384 & 384 & 384 &
    384 & 384 & 768 
    \\
    \textbf{MLP Hidden} & 384 & 384 & 384 &
    1536 & 1536 & 3072 
    \\
    \textbf{Grad Clip} & None & None & None &
    None & None & None 
    \\
    \bottomrule
\end{tabular}\\
\caption{Hyperparameter variations for various attention mechanisms (1 of 3).}
\end{table}

\begin{table}[h!]
\label{tab:hyperparam_variations2}
\centering
\begin{tabular}{c|cccccc}
    \toprule
    \textbf{Attention} & 
    Linear & Linear & Linear &
    DS & DS & DS\\
    \textbf{Dataset} & 
    C10 & C100 & Imagenette & 
    C10 & C100 & Imagenette\\
    \textbf{Init LR} & 
    1e-3 & 5e-5 & 1e-4 &
    4.5e-4 & 1e-3 & 5e-5
    \\
    \textbf{Batch Size} & 
    512 & 128 & 128 &
    256 & 128 & 128 
    \\
    \textbf{Dropout} &
    0.0 & 0.0 & 0.0 &
    0.0 & 0.0 & 0.0
    \\
    \textbf{Heads} & 
    12 & 12 & 12 &
    8 & 12 & 12 
    \\
    \textbf{Hidden} & 
    384 & 384 & 384 &
    384 & 384 & 768 
    \\
    \textbf{MLP Hidden} & 
    384 & 384 & 384 &
    384 & 384 & 3072 
    \\
    \textbf{Grad Clip} & 
    0.5 & 0.5 & 0.5 &
    None & None & None 
    \\
    \bottomrule
    \end{tabular}\\
    \caption{Hyperparameter variations for various attention mechanisms (2 of 3).}
\end{table}

\begin{table}[h!]
\centering
    \begin{tabular}{c|ccc}
        \toprule
        \textbf{Attention} &  
        Cosine & Cosine & Cosine\\
        \textbf{Dataset} & 
        C10 & C100 & Imagenette\\
        \textbf{Init LR} & 
        3e-4 & 1e-3 & 2e-4
        \\
        \textbf{Batch Size} & 
        512 & 128 & 128
        \\
        \textbf{Dropout} &
        0.0 & 0.0 & 0.0
        \\
        \textbf{Heads} & 
        8 & 12 & 12
        \\
        \textbf{Hidden} & 
        384 & 384 & 768
        \\
        \textbf{MLP Hidden} & 
        1024 & 384 & 384
        \\
        \textbf{Grad Clip} & 
        None & None & None
        \\
        \bottomrule
    \end{tabular}\\
    \caption{Hyperparameter variations for various attention mechanisms (3 of 3).}
    \label{tab:hyperparam_variations3}
\end{table}

\section{ Learning Curves }
This section provides learning curves for the validation accuracy of each attention mechanism under the scenario where both training and testing data were corrupted with fog. The curves illustrate the training dynamics and convergence behavior that led to the final performance metrics reported in the main paper.

We present four sets of learning curves. The first two show the absolute validation accuracy on CIFAR-10 and CIFAR-100, respectively, with the curves for all five attention mechanisms overlaid for direct comparison. The subsequent two graphs show the validation accuracy as a percentage of the clean-data baseline for CIFAR-10 and CIFAR-100, highlighting the relative robustness and performance degradation for each attention type throughout training.

It is important to note that the total number of training epochs varies between different attention mechanisms, as each trained at different speeds and required a different amount of training to reach convergence. However, for any single attention mechanism, the total number of training epochs was held constant across all corruption scenarios (clean, train-corrupted, test-corrupted, and both-corrupted) to ensure a fair comparison of its robustness.

\begin{figure}[!h]
    \centering
    \includegraphics[height=9cm,width=\linewidth,keepaspectratio]{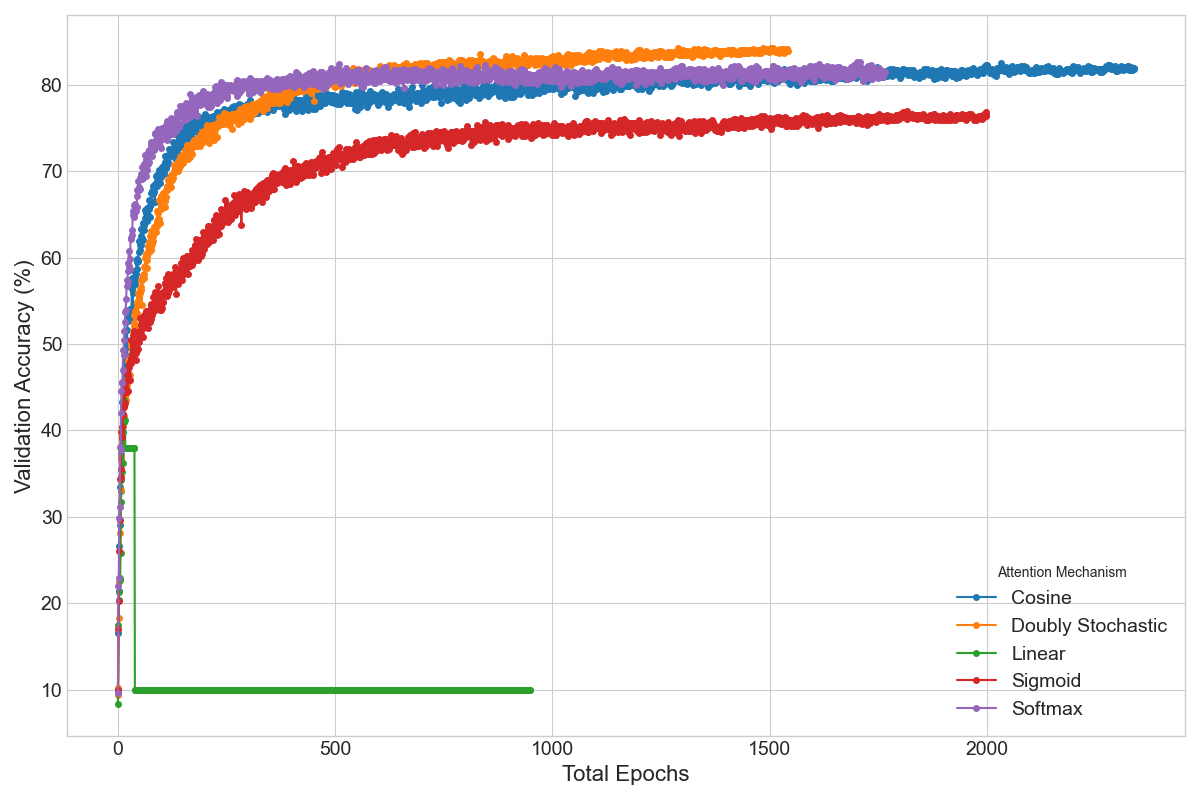}
    \caption{Validation accuracy learning curves for Vision Transformers using Cosine (\textcolor{blue}{blue}), Doubly Stochastic (\textcolor{orange}{orange}), Linear (\textcolor{green}{green}), Sigmoid (\textcolor{red}{red}), and Softmax (\textcolor{violet}{purple}) attention on the CIFAR-10 dataset where both training and test sets are corrupted. We observe that Doubly Stochastic attention not only trains stably but also achieves the highest final accuracy, demonstrating its superior robustness when learning from consistently corrupted data.}
    \label{fig:c10_absolute_training_curve}
\end{figure}
\begin{figure}[!h]
    \centering
    \includegraphics[height=9cm,width=\linewidth,keepaspectratio]{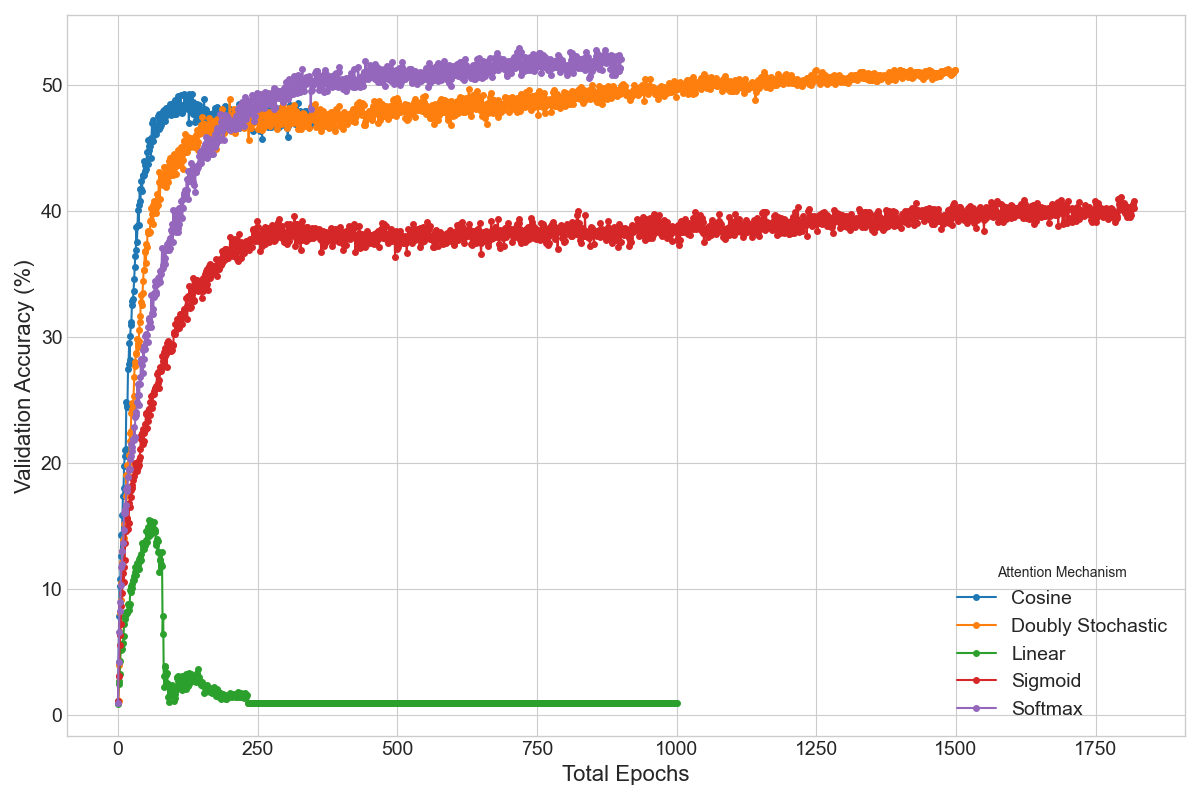}
    \caption{Absolute validation accuracy learning curves for Cosine (\textcolor{blue}{blue}), Doubly Stochastic (\textcolor{orange}{orange}), Linear (\textcolor{green}{green}), Sigmoid (\textcolor{red}{red}), and Softmax (\textcolor{violet}{purple}) attention on the more complex CIFAR-100 dataset with corrupted training and testing data. We observe that Softmax attention reaches the highest peak accuracy and Doubly Stochastic attention also demonstrates a stable and competitive learning trajectory.}
    \label{fig:c100_absolute_training_curve}
\end{figure}

\begin{figure}[!h]
    \centering
    \includegraphics[height=9cm,width=\linewidth,keepaspectratio]{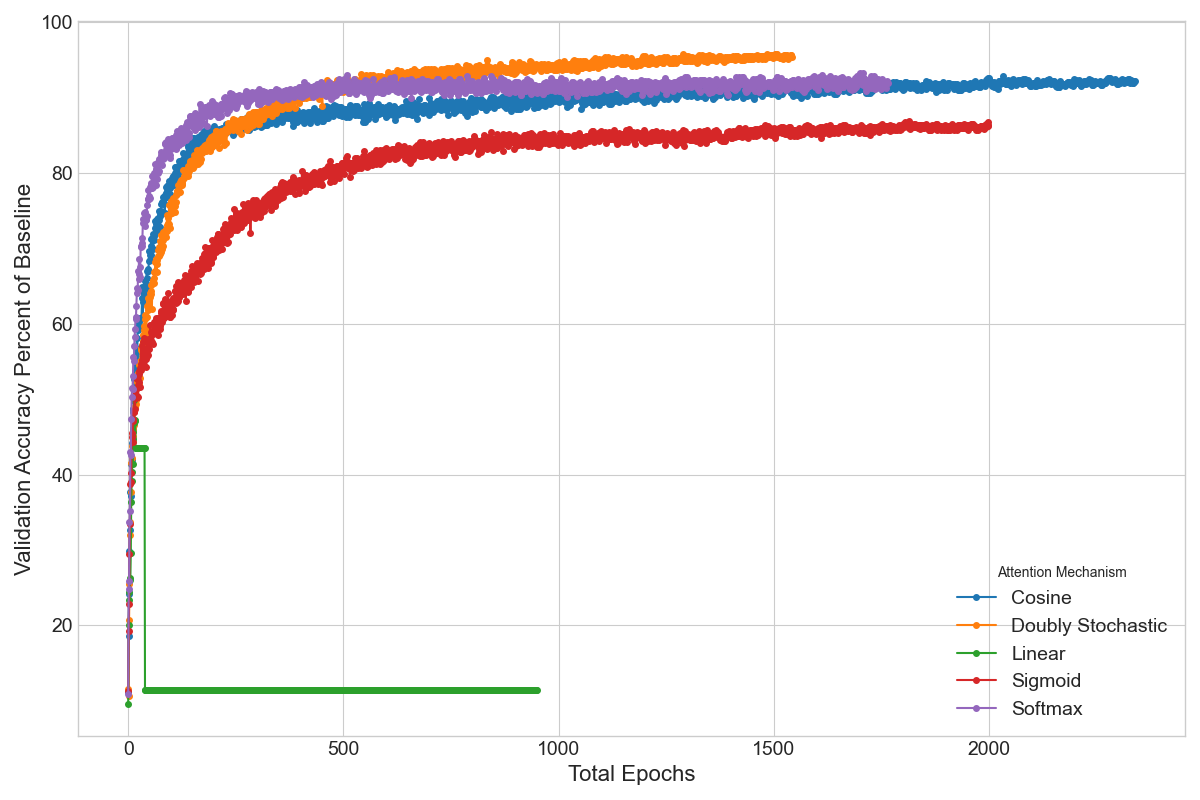}
    \caption{Relative validation accuracy (as a percentage of clean baseline) for Cosine (\textcolor{blue}{blue}), Doubly Stochastic (\textcolor{orange}{orange}), Linear (\textcolor{green}{green}), Sigmoid (\textcolor{red}{red}), and Softmax (\textcolor{violet}{purple}) attention on CIFAR-10 with corrupted training and testing data. We see clearly that Doubly Stochastic attention's performance degrades the least, as it consistently maintains the highest percentage of its original accuracy, confirming its superior resilience to corruption.}
    \label{fig:c10_absolute_training_curve}
\end{figure}
\begin{figure}[!h]
    \centering
    \includegraphics[height=9cm,width=\linewidth,keepaspectratio]{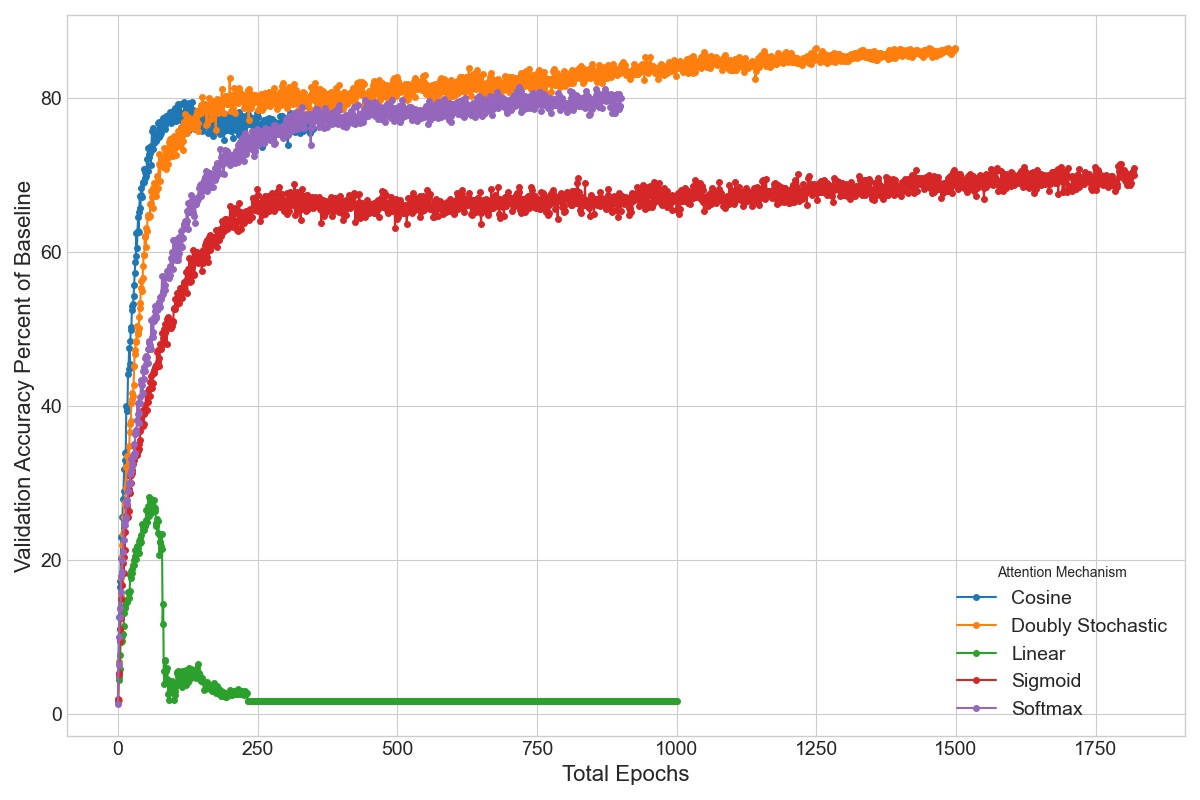}
    \caption{Relative validation accuracy learning curves for Cosine (\textcolor{blue}{blue}), Doubly Stochastic (\textcolor{orange}{orange}), Linear (\textcolor{green}{green}), Sigmoid (\textcolor{red}{red}), and Softmax (\textcolor{violet}{purple}) attention on the CIFAR-100 dataset with corrupted training and testing data. We find that Doubly Stochastic attention again demonstrates the strongest robustness, maintaining a significantly higher percentage of its baseline performance compared to all other mechanisms on this more challenging dataset. }
    \label{fig:c100_absolute_training_curve}
\end{figure}

\end{document}